\pdfoutput=1

\documentclass[11pt]{article}

\usepackage{emnlp2021}

\usepackage{times}
\usepackage{latexsym}

\usepackage[T1]{fontenc}

\usepackage[utf8]{inputenc}
\usepackage[T1]{fontenc}    
\usepackage{url}            
\usepackage{booktabs}       
\usepackage{amsfonts}       
\usepackage{nicefrac}       
\usepackage{microtype}      

\usepackage{graphicx}
\usepackage{subcaption}
\usepackage{enumitem}

\usepackage{csquotes}
\usepackage{todonotes}
\usepackage{xspace}
\usepackage{amsmath}
\usepackage{amssymb}
\usepackage{float}
\usepackage{wrapfig}
\usepackage{lipsum}
\usepackage{xcolor,colortbl}
\usepackage{fdsymbol}
\usepackage{multirow}
\usepackage{rotating}
\usepackage{algorithm,algpseudocode}

\newcommand{\code}[1]{\texttt{#1}}
\newcommand{\cmd}[1]{\textcolor{darkgreen}{\textbf{\small{\code{#1}}}}}

\usepackage{pifont}

\newcommand{\fone}{$\text{F}_1$\xspace}

\newcommand{\isquad}{iSQuAD\xspace}

\newcommand{\imrc}{iMRC\xspace}
\newcommand{\squad}{SQuAD\xspace}

\newcommand{\ctrlf}{Ctrl+F\xspace}
\newcommand{\query}{\textcolor{orange2}{\small{QUERY}}\xspace}

\definecolor{color1}{HTML}{da6752}
\definecolor{color2}{HTML}{5573a6}
\definecolor{green1}{HTML}{0b5400}
\definecolor{orange1}{HTML}{f3905c}
\definecolor{orange}{HTML}{ff7700}
\definecolor{cyan}{HTML}{008b8b}
\definecolor{blue}{HTML}{0000ff}
\definecolor{yellow1}{HTML}{e3ab12}
\definecolor{lb}{HTML}{deecff}
\definecolor{ly}{HTML}{fffeba}
\definecolor{purple1}{HTML}{9258cc}
\definecolor{blue1}{HTML}{027db5}
\definecolor{cyan1}{HTML}{00a0b5}
\definecolor{pink1}{HTML}{ff7a7a}
\definecolor{red1}{HTML}{8a0000}
\definecolor{blue2}{HTML}{041480}
\definecolor{yellowishgreen}{HTML}{2e8a00}
\definecolor{magenta}{HTML}{9b00a1}
\definecolor{darkblue}{HTML}{240394}
\definecolor{darkgreen}{HTML}{005e19}
\definecolor{orange2}{HTML}{de7c04}

\definecolor{lred}{HTML}{ffb8b8}
\definecolor{lgreen}{HTML}{b8ffc3}
\definecolor{lblue}{HTML}{b8e6ff}
\definecolor{lpurple}{HTML}{cab8ff}
\definecolor{lorange}{HTML}{ffdab8}
\definecolor{lcyan}{HTML}{abfff9}
\definecolor{lmagenta}{HTML}{ffbfed}
\definecolor{lightgray}{HTML}{e0e0e0}
\definecolor{lightyellow}{HTML}{feffde}

\newcommand{\clg}{\cellcolor{lightgray}}
\newcommand{\cly}{\cellcolor{lightyellow}}

\usepackage{array}
\newcolumntype{L}[1]{>{\raggedright\let\newline\\\arraybackslash\hspace{0pt}}m{#1}}
\newcolumntype{R}[1]{>{\raggedleft\let\newline\\\arraybackslash\hspace{0pt}}m{#1}}
\newcolumntype{C}[1]{>{\centering}m{#1}}

\usepackage{microtype}

\title{Interactive Machine Comprehension with Dynamic Knowledge Graphs}

\author{Xingdi Yuan\\
Microsoft Research, Montr\'{e}al \\
{\tt eric.yuan@microsoft.com }
}

\begin{document}
\maketitle
\begin{abstract}
Interactive machine reading comprehension (\imrc) is machine comprehension tasks where knowledge sources are partially observable. 
An agent must interact with an environment sequentially to gather necessary knowledge in order to answer a question.
We hypothesize that graph representations are good inductive biases, which can serve as an agent's memory mechanism in \imrc tasks.
We explore four different categories of graphs that can capture text information at various levels.
We describe methods that dynamically build and update these graphs during information gathering, as well as neural models to encode graph representations in RL agents.
Extensive experiments on \isquad suggest that graph representations can result in significant performance improvements for RL agents.\footnote{We release code and data at \url{https://github.com/xingdi-eric-yuan/imrc_graph_public}}

\end{abstract}

\section{Introduction}
\label{section:intro}

Machine reading comprehension (MRC) has gathered wide interest from the NLP community in recent years.
It serves as a way to benchmark a system's ability to understand and reason over natural language.
Typically, given a knowledge source such as a document, a model is required to read through the knowledge source to answer a question about some information contained therein.
In the extractive QA paradigm, in particular, answers are typically sub-strings of the knowledge source \citep{rajpurkar16squad,trischler16newsqa,yang18hotpot}.
Models are thus required to select a span from the knowledge source as their prediction.

A recent line of work known as interactive machine reading comprehension (\imrc) features interactive language learning and knowledge acquisition \citep{yuan2020imrc,ferguson2020iirc}.
It shifts the focus of MRC research towards a more realistic setting where the knowledge sources (environments) are \emph{partially observable}. 
Under this setting, agents must iteratively interact with the environment to discover necessary information in order to answer the questions.
The sequence of interactions between an agent and the environment may resemble the agent's reasoning path, rendering a higher level of interpretability in the agent's behaviour.
The trajectories of interactions can also be seen as procedural knowledge, which potentially brings agents extra generalizability (humans do not necessarily know the answer to a question immediately, but they know the procedure to search it).
Compared to many static MRC datasets, where the entire knowledge source (e.g., a paragraph in \squad) is presented to the model immediately, the \imrc setting may alleviate the risk of learning shallow pattern matching \citep{sugawara18easier,sen2020what}.

On a parallel track, there have been a plethora of studies that leverage graphs in MRC.
Multiple linguistic features have been explored to help construct graphs, such as coreference \citep{dhingra2018neural,song2018exploring}, entities \citep{de2018question,qiu2019dynamically,tu2020select} and semantic roles \citep{zheng2020srlgrn}. 
In related areas such as vision- and text-based games, prior works also attempt to build implicit and explicit graphs to encode data from various types of modalities \citep{johnson2016learning,ammanabrolu19graph,Kipf2020Contrastive,adhikari2020gata}.
All of these works, covering domains from static MRC to sequential decision making, suggest that graph representations can facilitate model learning. 
This gives us a strong motivation to leverage graph representations in the \imrc setting.

We hypothesize that graph representations are good inductive biases, since they can serve naturally as a memory mechanism to help RL agents tackle partial observability in \imrc tasks.
We develop an agent that can dynamically update new information into a graph representation at every step, the agent integrates information from both text and graph modalities to make decisions.
The main contributions of this work are as follows:
\begin{enumerate}
    \item We propose four categories of graph representations, each capturing text information from a unique perspective; we demonstrate how to generate and maintain the graphs dynamically in the \imrc tasks.
    \item We extend the RL agent proposed in \citep{yuan2020imrc} by adding a graph encoding mechanism and a recurrent memory mechanism.
    \item We conduct extensive experiments and show that the proposed graph representations can greatly boost agent's performance on \imrc tasks.
\end{enumerate}

\section{Problem Setting}
\label{section:problem_setting}

We follow the \imrc setting \citep{yuan2020imrc}, where given an environment consisting of a \emph{partially observable} document and a question, an agent needs to sequentially interact with the environment to discover necessary information and then answer the question. 
The \imrc paradigm reformulates existing MRC datasets (e.g., \squad) into interactive environments by occluding most parts of their documents.
A set of commands are defined to help agents reveal glimpses of the hidden documents.

\imrc can be seen as a controllable simulation to a family of complex real world environments, where knowledge sources are partially observable yet easily accessible by design through interactions.
One such example is the Internet, where humans can efficiently navigate through keywords and links to retrieve only the necessary information, rather than reading through the entire collection of websites.
While \imrc shares some common properties with multi-step retrieval \citep{yang18hotpot,zhao2021multi} and open-domain QA \citep{lewis2020question}, 
the focus here is to push the boundaries of information-seeking agents \citep{bachman16infoseeking} from an RL/navigation perspective.

Formally, an \imrc data-point (game) is a discrete-time, partially observable Markov decision process (POMDP) defined by $(S, T, A, \Omega, O, R, \gamma)$.
At game step $t$, the environment state $s_t \in S$ represents the semantics and information contained in the full document, as well as which subset of the sentences has been revealed to the agent.
The agent perceives text information as its observation, $o_t \in \Omega$, which depends on the environment state with probability $O(o_t|s_t)$. 
The agent issues an action $a_t \in A$, resulting in a state transition $s_{t+1}$ with probability $T(s_{t+1} | s_t, a_t)$ in the environment (i.e., a new sentence is shown to the agent).
Based on its actions, the agent receives rewards $r_t = R(s_t, a_t)$. 
The agent's objective is to maximize the expected discounted sum of rewards $E \left[\sum_t \gamma^t r_t \right]$, where $\gamma \in [0, 1]$ is the discount factor.

\paragraph{Difficulty Levels}
Given a question, only the first sentence of a document is initially exposed to an agent.
During information gathering phase, the agent uses the following commands to interact with the environment:
1) \cmd{previous} and 2) \cmd{next} will jump to the previous or next sentence, respectively; 
3) \cmd{\ctrlf} \query: jumps to the sentence with the next occurrence of \query; 4) \cmd{stop} terminates the interaction.
Whenever the agent issues the \cmd{stop} action, or it has exhausted its interaction budget \footnote{An agent can at most interact 20 steps.}, the information gathering phase is terminated, and the agent needs to answer the question immediately.
Thanks to the extractive nature of MRC datasets such as \squad, agents can label a span from its observation $o_t$ as prediction. 
Note that in order to correctly answer the question, an agent needs to effectively gather necessary information so that its observation $o_t$ contains the answer as a sub-string.

\citet{yuan2020imrc} define easy and hard as two difficulty levels.
In the easy mode, all four commands are available during information gathering phase; whereas in the hard mode, only \cmd{\ctrlf} and \cmd{stop} can be used.
Intuitively, in the easy mode, an agent can rely on the \cmd{next} command to traverse the entire document, which essentially reduces the problem to learning to stop at the right sentence.
In contrast, in the hard mode, the agent is forced to \cmd{\ctrlf} in a smart manner to navigate to potentially informative sentences.

\begin{figure*}[t!]
\centering
\begin{subfigure}{.5\textwidth}
  \centering
  \includegraphics[width=0.8\textwidth]{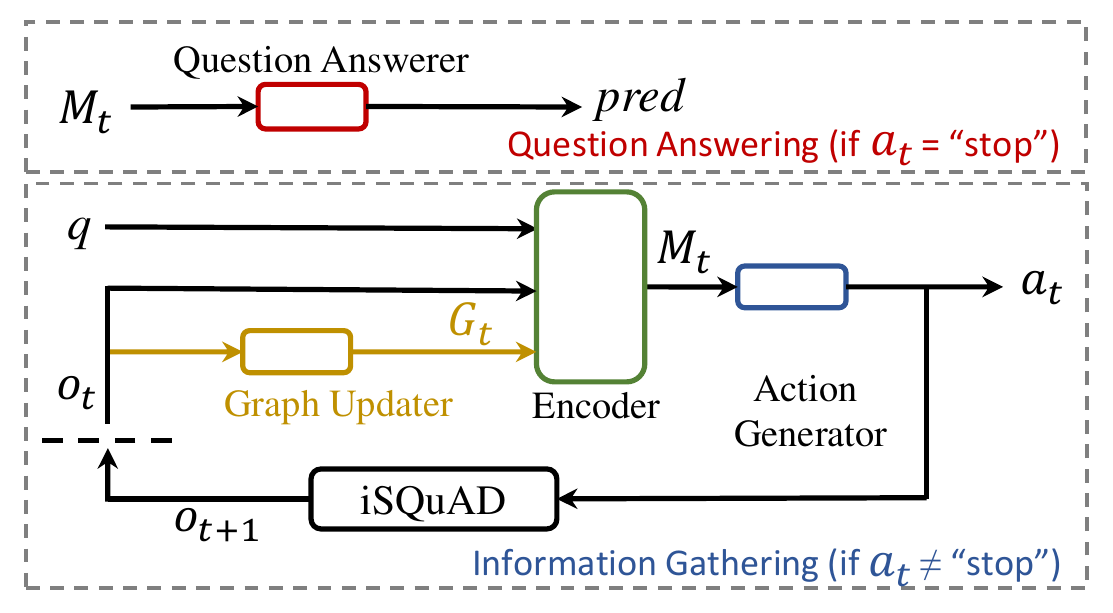}
\end{subfigure}%
\begin{subfigure}{.5\textwidth}
  \centering
  \includegraphics[width=0.9\textwidth]{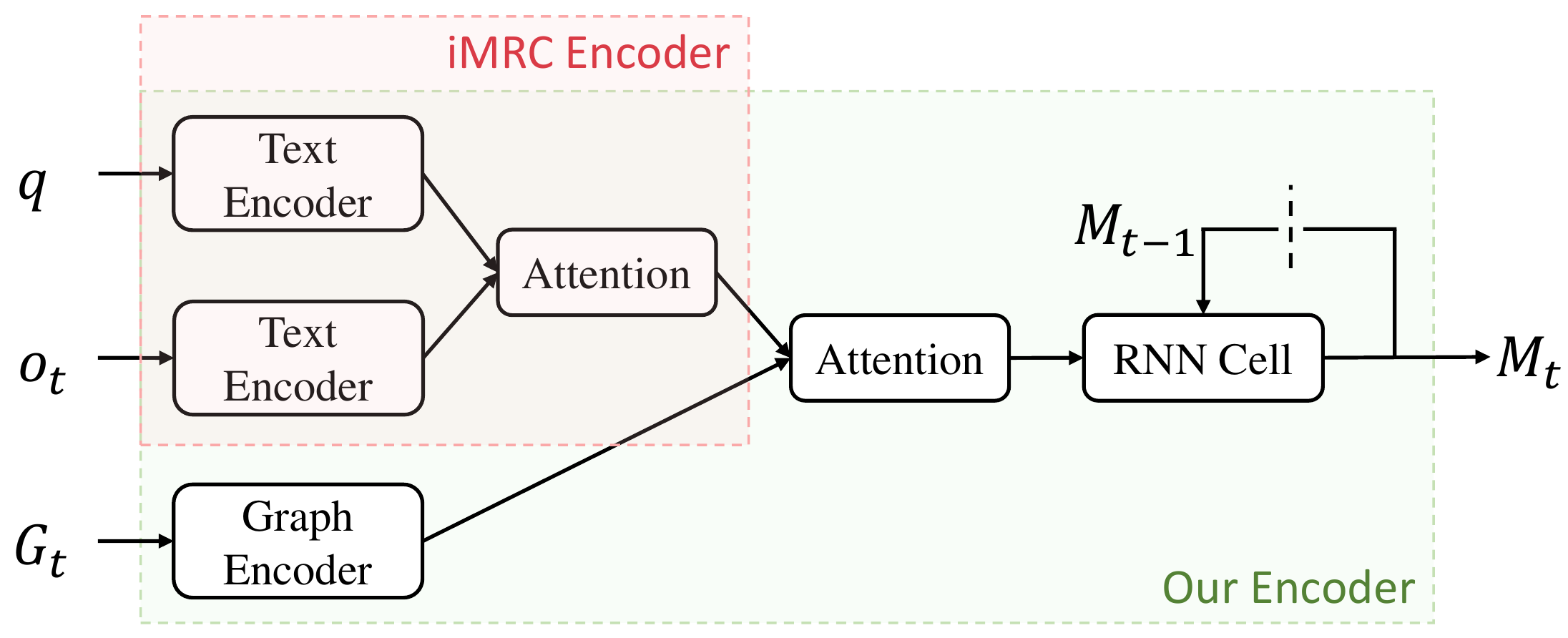}
\end{subfigure}
\caption{\textbf{Left:} an overview of our agent. We propose to use graph representations as an additional input modality to the \imrc agent \citep{yuan2020imrc}. \textbf{Right:} a zoomed in view of our encoder module, extended from \imrc.}
\label{fig:pipeline}
\end{figure*}

\paragraph{\query Types}
Three finer-grained settings are further defined according to the action space of the \cmd{\ctrlf} \query command.
Specifically, ranging from easy to hard, \query can be a token extracted from the question ($q$), the concatenation of the question and currently observable sentence ($q + o_t$), or any token selected from the dataset's vocabulary (\textit{vocab}).
From an RL perspective, the sizes of the three settings' action space can differ by several orders of magnitude (e.g., a question with 10 tokens in $q$ versus a vocabulary size of 20K in \textit{vocab}).

\section{Methodology}
\label{section:method}

In this work, we adopt the agent proposed in \citep{yuan2020imrc} as baseline, as shown in Figure~\ref{fig:pipeline}.
We propose to add a novel graph updater module and a graph encoding layer into the pipeline.
Specifically, at game step $t$, the graph updater takes the text observation $o_t$ and the graph $G_{t-1}$ from previous step as input and generates a new graph $G_t$.
Subsequently, the graph is encoded into hidden states, which is later aggregated with text representations. 
Note that distinct from fully observed Knowledge Graphs (KGs) in static MRC works, our graphs are dynamically generated, i.e., at every interaction step, our agent can update information from the new observation into its graph representations.


In this section, we will first introduce the key methods we use to generate and update the graph representations.
Later on, we will describe a Graph Neural Network (GNN)-based graph encoder which encodes the information carried by the graphs.
For the common components shared with \imrc, we refer readers to \citep{yuan2020imrc} or Appendix~\ref{appendix:agent_structure} for detailed information.

\paragraph{Notations}

We denote a graph generated at game step $t$ as $G_t=(\mathcal{V}_t, \mathcal{E}_t)$, where $\mathcal{V}_t$ and $\mathcal{E}_t$ represent the set of vertices (nodes) and edges (relations).
All graphs are directed by default. 
For two nodes $i \in \mathcal{V}$ and $j \in \mathcal{V}$, we denote the connection from $i$ to $j$ as $e_{i \rightarrow j} \in \mathcal{E}$.
We represent graph $G$ as an adjacency tensor, with the size of ${\mathcal{R} \times \mathcal{N} \times \mathcal{N}}$, where $\mathcal{R}$ and $\mathcal{N}$ denote the number of relations and nodes, respectively.
This tensor can either be binary or real-valued depending on graph type.

\subsection{Generating and Updating Graphs}
\label{subsection:graph_updater}

We propose four different graph representations.
The four graph types capture distinct aspects of information in text, from lower level linguistic features to high level semantics.

\subsubsection{Word Co-occurrence \textbf{\small{(Rule-based)}}}
\label{subsection:cooccur}
\begin{center} 
    \textcolor{blue1}{\textbf{\small{$G \in \{0, 1\}^{\mathcal{R} \times \mathcal{N} \times \mathcal{N}}, \mathcal{V}\text{: words}, \mathcal{E}\text{: sentences.}$}}}    
\end{center}

In word co-occurrence graphs, we connect tokens according to their co-occurrences. 
We assume that words appear in the same sentence tend to be relevant.
Common words across sentences further enable to build more complex graphs where each word is connected with multiple related concepts.

Omitting the notation of game step $t$ for simplicity, if two tokens $i \in \mathcal{V}$ and $j \in \mathcal{V}$ co-occur in a sentence $s$, the edge $e_{i \rightarrow j}$ between $i$ and $j$ is defined as $s$.
Computationally (e.g., for GNNs), the representations of $i$ and $j$ are their word embeddings, the representation of the relation $e_{i \rightarrow j}$ is the sentence encoding of $s$.
Note in this setting, graphs are symmetrical (i.e., $e_{i \rightarrow j} = e_{j \rightarrow i}$).
Typically, tokens $i$ and $j$ can co-occur in multiple sentences.
We thus allow multiple connections to appear between two graph nodes, each connection represents a particular sentence where they co-occur.

\subsubsection{Relative Position \textbf{\small{(Rule-based)}}}
\label{subsection:pos}
\begin{center}
    \textcolor{blue1}{\textbf{\small{$G \in \{0, 1\}^{\mathcal{R} \times \mathcal{N} \times \mathcal{N}}, \mathcal{V}\text{: words},$}}}\\
    \textcolor{blue1}{\textbf{\small{$\mathcal{E}\text{: relative position between words.}$}}}
\end{center}

In relative position graphs, we aim to capture and embed the word ordering and distance information.
This can be seen as capturing a loose form of the \textit{Subject-Verb-Object (SVO)} structure within sentences, without the need of parsing them.
The intuition of capturing token position information is supported by the idea of position embeddings in training large-scale language models such as BERT \citep{devlin2018bert,wang2021on}. 

In our setting, we first define a window size $l \in \mathbb{Z}^{+}$. 
For any two tokens $i$ and $j$ within a sentence $s$, their relation $e_{i \rightarrow j}$ is defined as:
\begin{equation}
\small
e_{i \rightarrow j} = 
    \begin{cases}
        l  & \text{if $\text{pos}_j - \text{pos}_i > l$,}\\
        -l & \text{elif $\text{pos}_j - \text{pos}_i < -l$,}\\
        \text{pos}_j - \text{pos}_i & \text{otherwise;}
    \end{cases}
\end{equation}
in which, $\text{pos}_i$ and $\text{pos}_j$ indicate the two tokens' position indices in $s$.
Therefore, the total number of relations $\mathcal{R} = 2l + 1$, the set of relations consists of all integers from $-l$ to $l$.
We also connect tokens with themselves via self-connections ($e_{i \rightarrow i} = 0$) to facilitate message passing in GNNs.

\subsubsection{Semantic Role Labeling \textbf{\small{(Parser-based)}}}
\label{subsection:srl}
\begin{center}
    \textcolor{blue1}{\textbf{\small{$G \in \{0, 1\}^{\mathcal{R} \times \mathcal{N} \times \mathcal{N}}, \mathcal{V}\text{: chunks returned by SRL},$}}}\\
    \textcolor{blue1}{\textbf{\small{$\mathcal{E}\text{: semantic role labels of the chunks.}$}}}    
\end{center}

\begin{figure}[t!]
    \centering
    \includegraphics[width=0.5\textwidth]{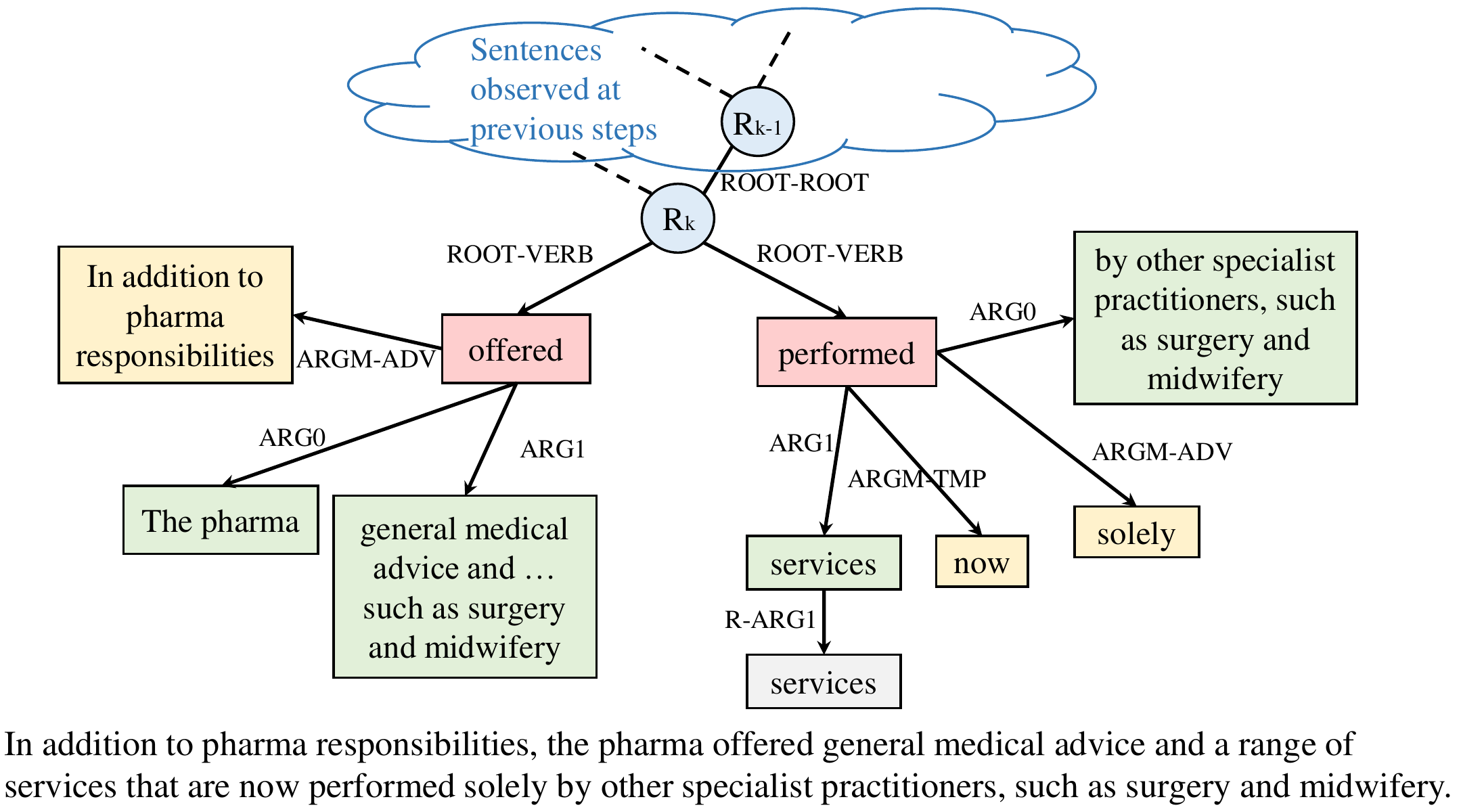}
    \caption{A partial view of the SRL graph corresponding to the given sentence. \textcolor{blue1}{Blue}: sentence root; \textcolor{red1}{red}: predicates/verbs; \textcolor{green1}{green}: arguments; \textcolor{yellow1}{yellow}: modifiers. }
    \label{fig:srl}
\end{figure}

Similar to recent approaches designed for static MRC tasks \cite{zheng2020srlgrn,zhang2020semantics}, we investigate building knowledge graphs via Semantic Role Labeling (SRL).
An SRL system \citep{shi2019simple} can detect the arguments associated with each of the predicates (or verbs) within a sentence and how they are classified into specific roles.
This property is essential in tackling MRC tasks, especially for extractive QA datasets, where answers are typically short chunks of text (e.g., entities), such chunks can often be identified as arguments by an SRL parser.
Via SRL, we can easily know how arguments interact with each other, further connecting common chunks detected in multiple sentences can produce an argument flow that helps understanding the paragraph.

In our SRL graphs\footnote{SRL is done with AllenNLP Toolkit (\url{https://demo.allennlp.org/semantic-role-labeling}).}, we use chunks that are identified as predicates, arguments, and modifiers as nodes. 
We use their semantic role labels w.r.t. their corresponding predicates as relations.
For longer sentences, an SRL system often returns multiple predicate-centric graphs.
We define a special \cmd{ROOT} node for each sentence, and connect it with all the predicates within the sentence, with a special relation \cmd{ROOT-VERB}.
We connect sentences by connecting their root nodes using a relation \cmd{ROOT-ROOT}.
An example of the SRL graph is shown in Figure~\ref{fig:srl}.

In order to facilitate models such as GNNs (e.g., easier message passing, denser signals), we define a set of reversed relations in SRL graphs.
For instance, in Figure~\ref{fig:srl}, $e_{\text{performed} \rightarrow \text{services}}$ is \cmd{ARG1}, we use an \cmd{ARG1-rev} relation as $e_{\text{services} \rightarrow \text{performed}}$ in our experiments.
However, we ignore the reversed relations in Figure~\ref{fig:srl} for simplicity.

\subsubsection{Continuous Belief \textbf{\small{(Trainable model)}}}
\label{subsection:gata}
\begin{center}
    \textcolor{blue1}{\textbf{\small{$G \in [-1, 1]^{\mathcal{R} \times \mathcal{N} \times \mathcal{N}}, \mathcal{V}\text{: concepts determined by agent},$}}}\\
    \textcolor{blue1}{\textbf{\small{$\mathcal{E}\text{: relations determined by agent.}$}}}
\end{center}

In addition to rule- and parser-based graph updaters, we also investigate a data-driven approach.
Inspired by \citet{adhikari2020gata}, we use self-supervised learning technique to pre-train a graph updater that can maintain a \emph{continuous} belief graph for the \imrc task.
Specifically, given the new observation $o_t$, we use a neural network to modify graph $G_{t-1}$ from previous step, to produce a new graph $G_t$.
Without the need of manually defining and hand-crafting specific graph structures --- which may inject unnecessary prior into the agent ---
we assume that as long as a learned graph $G_t$ can be used to reconstruct $o_t$, the graph should have contained useful information of the text.

However, learning to reconstruct $o_t$ in a word-by-word fashion requires the model to learn features that are less useful in the \imrc task.
Therefore, we adopt a contrastive representation learning strategy to approximate the reconstruction. 
Specifically, we train a discriminator $D$ that differentiates between true $o_t$ (positive samples) and a ``corrupted'' version of them $\widetilde{o}_t$ (negative samples), conditioned on $G_t$.
This relieves the model of the burden to learn syntactical features (NLG), so that it can focus more on the semantic side instead.
We use a standard noise-contrastive objective to minimize the binary cross-entropy (BCE) loss \citep{velickovic2018deep}:
\begin{equation}
    \small
    \begin{aligned}
        \mathcal{L} = \frac{1}{K} \textstyle\sum_{t=1}^{K} (
            &\mathbb{E}_{o} [ \text{log} D( h_{o_t}, h_{G_t} ) ]  +\\
            &\mathbb{E}_{\widetilde{o}} [ \text{log} (1 - D( h_{\widetilde{o}_t}, h_{G_t} ) ) ] ),
    \end{aligned}
\end{equation}
where $K$ is the number of sentences in a \squad paragraph. 
To facilitate this pre-training, we utilize an online Wikipedia dump \citep{wilson_wiki}.
We remove all articles that appear in the \squad dataset, and use the rest as our negative sample collection \footnote{We apply the filtering to prevent the pre-trained graph updater from ``memorizing'' text that may appear in \squad validation set}.
The graph updater is then trained to generate graphs $G_t$ that can be used to differentiate between 1) sentences within current document and 2) sentences sampled from another Wikipedia article.

Note that $G_t$ is not explicitly grounded to any ground-truth graphs. Instead, they are essentially latent recurrent state representations, encoding information the agent has seen so far.
Therefore, the nodes and relations in these graphs are determined by the agent itself in a data-driven manner.
As a result, the adjacency tensors in these graphs are real-valued.
We provide more details of this graph updater in Appendix~\ref{appendix:gata_updater}.

\subsection{Encoding Graph Representations}
\label{subsection:graph_encoder}

We adopt a multi-layer relational graph convolutional network (R-GCN) \citep{schlichtkrull2018rgcn,adhikari2020gata} as our graph encoder.
Specifically, at the $l$-th layer of the R-GCN, for each node $i \in \mathcal{V}$, given the set of its neighbor nodes $\mathcal{V}^e_i \in \mathcal{V}$ under relation $e \in \mathcal{E}$, the R-GCN computes:
\begin{equation}
    \small
    \tilde{h}_i = \sigma\left(\sum_{e\in \mathcal{E}}\sum_{j\in \mathcal{V}^e_i} W^{l}_e [h^{l}_j; \text{Emb}_e] + W^{l}_0 [h^{l}_i; \text{Emb}_e]\right),
\end{equation}
where $W^{l}_e$ and $W^{l}_0$ are trainable parameters.
When the graph is discrete (i.e., word co-occurrence, relative position, SRL), we use ReLU as the activation function $\sigma$; when the graph is continuous (i.e., continuous belief), we use Tanh function as $\sigma$ to stabilize the model.

When the labels of graph nodes consist of tokens, we integrate their word representations into graph computation.
Specifically, for a node $i$, we use the concatenation of a randomly initialized node embedding vector and the averaged word embeddings of node label as the initial input $h_i^{0}$.
Similarly, for each relation $e$, $\text{Emb}_e$ is the concatenation of a randomly initialized relation embedding vector and the averaged word embeddings of $e$'s label.

We utilize highway connections \citep{srivastava15highway} between R-GCN layers:
\begin{equation}
\small
\begin{aligned}
    g &= \text{Sigmoid}(W_{\text{hw}}(\tilde{h}_i)), \\
    h^{l+1}_i &= g \odot \tilde{h}_i + (1 - g) \odot h^{l}_i, \\
\end{aligned}
\end{equation}
where $\odot$ indicates element-wise multiplication, $W_{\text{hw}}$ is a linear layer.
We denote the final output of the R-GCN as $h_{G_t} \in \mathbb{R}^{\mathcal{N} \times H}$, where $\mathcal{N}$ is the number of nodes in the graph, $H$ is hyperparameter.

\subsection{Aggregating Multiple Modalities}
\label{subsection:aggregate_modalities}

Following \citet{yuan2020imrc}, we utilize the context-query attention mechanism \citep{yu18qanet} to aggregate multiple representations. 
The inputs to a context-query attention layer are typically two sequences of representation vectors (e.g., sequence of tokens for text, sequence of nodes for graphs).
The attention computes element-wise similarity scores between the two inputs, then each element in one input can be represented by the weighted sum of the other input, and vice versa.

As shown in Figure~\ref{fig:pipeline} (right), we stack another context-query attention layer on top of the encoder used in \imrc, to aggregate the text representation (which encodes information in $o_t$ and $q$) with graph representation.
We denote the output from the second attention layer as $h_{og} \in \mathbb{R}^{L_{o_t} \times H}$, where $L_{o_t}$ is the length of $o_t$, $H$ is hyperparameter.

Although all the four graph types we investigate are updated dynamically, they can only represent the agent's belief of the \emph{current} state $s_t$.
There are clearly some information hard to be represented in the graphs, such as how did an agent navigate to the current sentence (i.e., the trajectories).
We thus leverage a recurrent neural network to incorporate history information into encoder's output representations.
Specifically, we use a GRU \citep{cho2014gru} as the recurrent component:
\begin{equation}
\small
\begin{aligned}
    h_{\text{inp}} &= \textrm{MaskedMean}(h_{og}),\\
    M_t &= \mathrm{GRU}(h_{\text{inp}}, M_{t-1}),
\end{aligned}
\end{equation}
in which, $h_{\text{inp}} \in \mathbb{R}^{H}$. $M_{t-1}$ is the output of the GRU cell at game step $t-1$.
As shown in Figure~\ref{fig:pipeline} (left), the output of encoder, $M_t$, is then used to both generating actions during information gathering phase, as well as extracting answers during question answering phase, this procedure exactly follows the \imrc pipeline.

\begin{table*}[t!]
    \centering
    \small
    \begin{tabular}{c|c|ccc|ccc|c|c}
        \toprule
        && \multicolumn{3}{c|}{Easy Mode} & \multicolumn{3}{c|}{Hard Mode} & & \\
        \#Mem Slot&Agent & $q$ & $q + o_t$ & \textit{vocab} & $q$ & $q + o_t$ & \textit{vocab} & \%RI & Reference\\
        \hline
        \multirow{6}{*}{1} & \clg\imrc \citep{yuan2020imrc}      &  \clg0.575 & \clg0.579 & \clg0.583 & \clg0.524 & \clg0.357 & \clg\textbf{0.264} & \clg-- & \multirow{6}{*}{\S~\ref{subsection:a1}.A1}\\
        &Ours (co-occur)          &  \cly0.632 & \cly0.624 & \cly0.635 & \cly0.582 & \cly0.426 & 0.258 & \cly9.16 & \\
        &Ours (rel. pos.)         &  \cly\textbf{0.634} & \cly0.634 & \cly\textbf{0.642} & \cly0.562 & \cly\textbf{0.440} & 0.250 & \cly9.18 & \\
        &Ours (SRL)               &  \cly0.616 & \cly\textbf{0.641} & \cly0.638 & \cly\textbf{0.603} & \cly0.434 & 0.253 & \cly\textbf{9.98} & \\
        &Ours (cont.)             &  \cly0.617 & \cly0.628 & \cly0.616 & \cly0.597 & \cly0.436 & 0.257 & \cly9.14 & \\
        &Ours (\textcolor{darkgreen}{ensemble})          &  \cly0.677 & \cly0.691 & \cly0.686 & \cly0.627 & \cly0.472 & \cly0.276 & \cly18.53 & \\
        \hline
        \multirow{6}{*}{3} &\clg\imrc \citep{yuan2020imrc}      &  \clg0.637 & \clg0.651 & \clg0.624 & \clg0.524 & \clg0.362 & \clg0.261 & \clg-- & \multirow{6}{*}{\S~\ref{subsection:a2}.A2}\\
        &Ours (co-occur)          &  \cly0.674 & \cly0.665 & \cly\textbf{0.675} & \cly0.605 & \cly\textbf{0.446} & 0.260 & \cly\textbf{9.06} & \\
        &Ours (rel. pos.)         &  \cly0.677 & \cly0.665 & \cly0.664 & \cly\textbf{0.615} & \cly0.438 & 0.257 & \cly8.66 & \\
        &Ours (SRL)               &  \cly\textbf{0.681} & \cly\textbf{0.678} & \cly0.654 & \cly0.600 & \cly0.440 & 0.258 & \cly8.45 & \\
        &Ours (cont.)             &  \cly0.676 & \cly0.642 & \cly0.662 & \cly0.592 & \cly0.426 & \cly\textbf{0.282} & \cly8.26 & \\
        & Ours (\textcolor{darkgreen}{ensemble})         &  \cly0.714 & \cly0.713 & \cly0.701 & \cly0.650 & \cly0.471 & \cly0.278 & \cly15.80 & \\
        \hline
        \multirow{6}{*}{5} &\clg\imrc \citep{yuan2020imrc}      &  \clg0.666 & \clg0.656 & \clg0.661 & \clg0.551 & \clg0.364 & \clg0.218 & \clg-- & \multirow{6}{*}{\S~\ref{subsection:a2}.A2}\\
        &Ours (co-occur)          &  \cly0.680 & \cly0.670 & \cly0.665 & \cly0.628 & \cly0.444 & \cly\textbf{0.258} & \cly9.84 & \\
        &Ours (rel. pos.)         &  \cly0.686 & \cly0.677 & \cly0.665 & \cly0.622 & \cly0.446 & \cly0.253 & \cly9.76 & \\
        &Ours (SRL)               &  \cly0.675 & \cly0.680 & \cly0.680 & \cly0.609 & \cly0.441 & \cly0.257 & \cly9.56 & \\
        &Ours (cont.)             &  \cly\textbf{0.699} & \cly\textbf{0.693} & \cly\textbf{0.696} & \cly\textbf{0.629} & \cly\textbf{0.455} & \cly0.257 & \cly\textbf{12.19} &  \\
        & Ours (\textcolor{darkgreen}{ensemble})         &  \cly0.736 & \cly0.733 & \cly0.725 & \cly0.665 & \cly0.484 & \cly0.277 & \cly18.79 & \\
        
        \bottomrule
    \end{tabular}
    \caption{Testing \fone scores and the relative improvement \%RI (averaged over six settings in a row). Best \textbf{single agent} scores within each setting are highlighted with \textbf{boldface}, scores better than \imrc are shaded in yellow.
    }
    \label{tab:result_a1a2}
\end{table*}

\section{Experiments and Results}
\label{section:exp_and_result}


We conduct experiments on the \isquad dataset \citep{yuan2020imrc} to answer three key questions:
\begin{itemize}
    \item Q1: Do graph representations help agents achieving better performance? In particular, among the four graph types, which of them provides the most performance boost?
    \item Q2: Do graph representations remain helpful in settings where multiple memory slots (observation queues) are available?
    \item Q3: If graph representations are great, can we get rid of the text modality?
\end{itemize}

\textbf{Experiment Setup:}
The \isquad dataset \citep{yuan2020imrc} is an interactive version of the \squad dataset \citep{rajpurkar16squad}, which consists of 82k/5k/10k environments for training, validation, and testing.
As described in Section~\ref{section:problem_setting}, \isquad contains two difficulty levels and three finer-grained \query type settings, all of which influence an RL agent's action space.
The environment provides an observation queue with $k$ memory slots depending on different configurations, where $k \in \{1, 3, 5\}$. 
The observation queue stores the $k$ most recent observation sentences to alleviate difficulties caused by partial observability.
Note in configuration where $k=1$, there is no history information stored.

Inherited from the original \squad dataset, an agent is evaluated by the $F_1$ score between its predicted answer and the ground-truth answers.
We compare our agents equipped with graph representations against \imrc scores reported in \citep{yuan2020imrc}, specially, we also report an agent m's relative improvement over \imrc:
\begin{equation}
    \small
    \%\text{RI} = (\text{F}_{1}^{\text{m}} - \text{F}_{1}^{\text{\imrc}}) / \text{F}_{1}^{\text{\imrc}} \times 100.0.
\end{equation}
For all experiment settings, we train the agent with three different random seeds.
We compute an agent's test score using the model checkpoint that achieves the best validation score.



\subsection*{A1: Graph representations indeed help, and ensemble is an useful strategy.}
\label{subsection:a1}

Intuitively, a dynamically maintained graph can serve as an agent's episodic memory. 
Therefore, the less information is provided by the environment, the more useful the graphs can be.
We first investigate our graph aided agent's performance on the game configuration where only single memory slot is available.
This is arguably the most difficult configuration in \imrc, where any valid action can lead to a completely different observation (a new sentence).
As a result, agents needs to rely on its own memory mechanism.

As shown in Table~\ref{tab:result_a1a2} (\#Mem Slot = 1), our agent outperforms \imrc in most of the settings by a noticeable margin.
We observe that the improvement brought by graph representations is consistent across the four graph types.
All of the four graph types provide over 9\% of average relative improvement over \imrc.
Among the four graph types, relative position graph and SRL graph seem to show advantage over the other two types, but this trend is not as significant.

Following standard strategy of model ensemble in MRC works, we test the ensemble of the four graphs.
Specifically, taking four individual agents, each trained with its corresponding graph types, we mix their decisions during test.
During the information gathering phase, we sum up the four agents' output probabilities, including the probabilities over action words (i.e., \cmd{previous}, \cmd{next}, \cmd{ctrl+f}, \cmd{stop}) and the probabilities over the \query tokens.
The four agents consequently take the action with the max summed probabilities to keep interacting with the environment.
During the question answering phase, we also sum up the output probabilities (over tokens in the sentence where the agents stop), and generate answers accordingly.
Surprisingly, we find that the ensemble greatly boosts agent's performance.
As shown in Table~\ref{tab:result_a1a2} (\#Mem Slot = 1), the ensemble agent nearly doubles our agent's relative improvement over \imrc. 
It is also worth noting that with ensembling, our agent achieved better score than \imrc in the Hard Mode + \textit{vocab} setting, which all the individual agent fail to outperform the baseline.
This observation aligns with our motivation that the four types of graphs capture different aspects of the information and thus may be complementary to each other.

\begin{table}[t!]
    \centering
    \small
    \begin{tabular}{ccccc}
        \toprule
        \multicolumn{5}{c}{Text Only \citep{yuan2020imrc}}\\
        \multicolumn{5}{c}{0.575}\\
        \hline
        \multicolumn{5}{c}{Graph Only}\\
        co-occur & rel. pos. & SRL & cont. & \textcolor{darkgreen}{ensemble}\\
        0.543 & 0.528 & 0.398 & 0.308 & 0.534\\
        \hline
        \multicolumn{5}{c}{Text + Graph}\\
        co-occur & rel. pos. & SRL & cont. & \textcolor{darkgreen}{ensemble}\\
        0.632 & 0.634 & 0.616 & 0.617 & 0.677 \\
        \bottomrule
    \end{tabular}
    \caption{Testing \fone  with different input modalities. \\Reference: \S~\ref{subsection:a3}.A3.}
    \label{tab:result_a3}
\end{table}

\subsection*{A2: Graph representations remain helpful even with explicit memories.}
\label{subsection:a2}

As mentioned above, in some configurations, the \isquad environment provides an observation queue that caches most recent few observations as an explicit memory mechanism.
A natural thing to explore is that if the advantages of equipping graph representations tend to diminish when the partial observability of the environments decreases. 

We train and test our agent using \isquad's configurations where 3 or 5 memory slots are available (i.e., at game step $t$, the input $o_t$ to the agent is the concatenation of the most recent 3 or 5 sentences it has seen).
From Table~\ref{tab:result_a1a2}, we observe that the previously observed trends are consistent across different memory slot number configurations.
Particularly, in the settings with 3 or 5 memory slots, our single agents equipped with graph representations can outperform \imrc in most of the settings.
All graph types provide a greater than \%8 of averaged relative improvements.
Again, the ensemble agent nearly doubles single agents' relative improvement over \imrc.
In the setting with 5 memory slots, we observe that the continuous belief graph is consistently outperforming its counterparts, which provides a \%12.19 of relative improvement.

Given the observation that graph representations seem still helpful even with explicit memories, we further compare graph as memory mechanism against the explicit memory slots provided by \isquad environments.
Comparing our best graph aided agent (receiving single sentence as input) against \imrc (receiving 3 and 5 sentences as input), we find our agent achieves a \%13.03 and \%13.47 of relative improvements over \imrc.
This suggests that the design of the memory mechanism plays a big role in the interactive reading comprehension tasks.
Although the concatenation of memory slots may provide as much amount of information, the inductive bias of graph representations are stronger.

\subsection*{A3: Text modality is necessary.}
\label{subsection:a3}

Based on our findings in previous subsections, we further investigate whether the dynamic graphs can replace the text modality.
We conduct a set of ablation experiments on the Easy Mode + $q$ games, with single memory slot.
Specifically, at every game step $t$, given the new observation $o_t$, we use the graph updater (described in Section~\ref{subsection:graph_updater}) to generate graph representations $G_t$.
Encoded by the graph encoder (described in Section~\ref{subsection:graph_encoder}), we directly aggregate the graph encoding with the question representations for further computations. 
In this way, the observation sentence $o_t$ is only used to build the graph, without serving as a direct input modality to the agent. \footnote{Note $o_t$ is absent only during the information gathering phase, due to the extractive design of the question answerer.}

After training and testing such graph-only variants of our agent, we compare them against the text-only version \citep{yuan2020imrc} and our full agent with both input modalities in Table~\ref{tab:result_a3}.
We observe that the graph-only agent fails to outperform the text-only baseline with any of the graph types, even with the ensemble of them.
This suggests that even though the text and graph modalities may contain redundant information (because the graphs are generated from the text), they represent the information complementarily in some sense.
We suspect the attention mechanism integrating text representations and graph representations (described in Section~\ref{subsection:aggregate_modalities}) may have contributed to the improvement of the full agent.
For instance, the agent may have learned to focus on certain sub-graph conditioned on tokens in $o_t$, and vice versa. 

\begin{table}[t!]
    \centering
    \small
    \begin{tabular}{C{0.7cm}|c|C{0.5cm}C{0.4cm}C{0.4cm}C{0.4cm}c}
        \toprule
        \#Mem & Agent & co-occur & rel. pos. & SRL & cont. & \textcolor{darkgreen}{ensemble}\\
        \hline
        \multirow{2}{*}{1} & Ours &         9.16 & 9.18 & 9.98 & 9.14 & 18.53 \\
        & w/o RNN &                         4.73 & 4.27 & 5.50 & 5.23 & 12.29 \\
        \hline
        \multirow{2}{*}{3} & Ours &         9.06 & 8.66 & 8.45 & 8.26 & 15.80 \\
        & w/o RNN &                         5.89 & 7.48 & 2.91 & 5.43 & 14.15 \\
        \hline
        \multirow{2}{*}{5} & Ours &         9.84 & 9.76 & 9.56 & 12.19 & 18.79 \\
        & w/o RNN &                         6.51 & 5.97 & 6.29 & 6.02 & 14.99 \\
        \bottomrule
    \end{tabular}
    \caption{Averaged \%RI over \imrc, comparing full agent with variants without RNN in encoder. Reference: \S~\ref{subsection:ablation}.Additional Results.}
    \label{tab:ablation}
\end{table}

\subsection*{Additional Results and Discussion}
\label{subsection:ablation}

As described in Section~\ref{subsection:aggregate_modalities}, our agent utilizes a recurrent component to be aware of history information, this component is absent in the original \imrc architecture.
Therefore, it is important to make sure the performance improvement shown in previous subsections are not solely caused by the RNN.
We conduct a set of experiments, with the RNN layer disabled (i.e., the output of attention layer becomes $M_t$ in Figure~\ref{fig:pipeline} right).

We show results of these experiments in Table~\ref{tab:ablation}, due to space limitation, we only show the averaged relative improvement over \imrc, readers can find full results in Appendix~\ref{appendix:full_results}.
Overall, graph representations contribute more to the improvement (for single agents, more than \%5 on average), this is especially clear for the ensemble agents, where even without RNN, agents can sometimes achieve very close performance with the full agent.
However, the effect of the RNN is non-negligible.
This again emphasizes the importance of memory mechanism in interactive reading comprehension tasks.
From our finding, multiple distinct memory mechanisms (i.e., memory slots, graphs, RNN cells) do not seem redundant, rather, they work cooperatively to produce a better score than solely using any of them.

It is noticeable in Table~\ref{tab:result_a1a2} that all agents performs poorly on Hard Mode + \textit{vocab} games.
This reveals limitations of RL-based algorithm (such as deep Q-learning we use in this work) --- when the action space is extremely large, the agent has near-zero probability to experience a trajectory that leads to any positive reward, and thus struggles to learn useful strategies.
This can potentially be mitigated by pre-training the agent with an easier setting then fine-tune in the difficult setting so that the agent has higher probability to experience good trajectories to start with.

A recent work \citep{guo2021text} propose to facilitate RL learning in tasks with huge action spaces (e.g., natural language generation) using Path Consistency Learning (PCL). 
Their PCL-based training method can update Q-values of all actions (tokens in vocabulary) at once, as opposed to only update the selected action (one token) in vanilla Q-Learning.
This can potentially enable \imrc agents to perform in a more natural and generic manner, for instance, to \cmd{\ctrlf} multi-word expressions as \query.

Due to space limitation, we report detailed agent structure, more results, and implementation details in Appendices.

\section{Related Work}
\label{section:related_work}

MRC has become an ever-growing area in the past decade, especially since the success of deep neural models.
Like an adversarial game, researchers release new datasets \citep{hill2015goldilocks,chen2016thorough,rajpurkar16squad,trischler16newsqa,nguyen16msmarco,reddy18coqa,yang18hotpot,choi18quac,tydiqa} and novel models \citep{trischler2016natural,wang16matchlstm,seo2016bidirectional,wang2017gated,huang2018fusionnet} one after another.
Since the flourishing of large scale pre-trained language models such as BERT~\citep{devlin2019bert}, RoBERTa~\citep{liu2019roberta} and XLNet~\citep{yang2019xlnet}, performance of neural models on MRC datasets have improved greatly.

While some researchers believe models have achieved human-level performance, others argue that there have been biases or trivial cues injected into MRC datasets unconsciously \citep{agrawal2016analyzing,weissenborn2017making,mudrakarta2018model,sugawara18easier,niven2019probing,sen2020what}.
These biases may cause models to learn shallow pattern matching, rather than deep understanding and reasoning skills.

\imrc \citep{yuan2020imrc} is a line of research that assumes partial observability and insufficient information.
To answer a question, models have to actively collect necessary information by interacting with the environment.
The \imrc paradigm can be described naturally within the RL framework, and thus it shares interests with video games \citep{badia20agent57}, text-based games \citep{ammanabrolu19graph,adhikari2020gata} and navigation \citep{mattersim,shridhar2020alfred}.

Graph construction is also a thriving direction lies at the intersection of multiple areas such as information extraction \citep{angeli2015leveraging}, knowledge base \citep{shin2015incremental}, logical reasoning \citep{sinha2019clutrr} and representation learning \citep{Kipf2020Contrastive}. 
Leveraging automatically constructed graph representations in static MRC has been shown effective, researchers use a wide range of linguistic features to help constructing graphs.
\citet{dhingra2018neural,song2018exploring} build graphs use coreference relations,
\citet{de2018question,qiu2019dynamically,tu2020select} leverage mentions of entities and
\citet{zheng2020srlgrn} build SRL graphs using parsers. 
In the context of RL, prior work have also shown that constructing graph representations from other modalities can be helpful to solve tasks in interactive environments \citep{johnson2016learning,Yang2018GLoMoUL,ammanabrolu19graph,adhikari2020gata}.

\section{Broader Impact}
\label{section:impact}

Our work is a proof-of-concept study, we use a relatively simple and restricted (in terms of both observations and actions) QA dataset, \isquad, for both training and evaluation. 
Although the current version of our work might have limited consequences for society, we believe that taking a broader view of our work can be beneficial by preventing our future research from causing potential social and ethical concerns.

Similar to many RL-based systems, the information gathering module of our agent is optimized solely on its performance w.r.t. the final metric, without much constraints on its behavior at each game step.
This can potentially make the system vulnerable since the RL agent may develop undesirable strategies that optimize the final metric.

In our current setting, the action space of the information gathering module is restricted (see Section~\ref{section:problem_setting}). 
However, if we consider a more general setting, e.g., to equip the agent with a larger action space by allowing it to generate a sequence of tokens as the \query to the \cmd{\ctrlf} action, we have to be extra careful about the aforementioned side effects caused by RL training.
For instance, the agent may develop unfavorable behaviors such as forgetting proper syntax, abusing certain pronouns, to optimize its final rewards.

\section{Conclusion}
\label{section:conclusion}
We explore to leverage graph representations in the challenging \imrc tasks.
We investigate different categories of graph structures that can capture text information at various levels.
We describe methods that dynamically generate the graphs during information gathering.
Experiment results show that graph representations provide consistent improvement across settings. 
This evinces our hypothesis that graph representations are proper inductive biases in \imrc.

\section*{Acknowledgments}
We thank Marc-Alexandre C\^ot\'{e}, 
Jie Fu and
Tong Wang for the helpful discussions about this work.
We also thank the anonymous EMNLP reviewers and area chairs for their helpful feedback and suggestions.

\bibliographystyle{acl_natbib}
\bibliography{biblio}

\clearpage
\appendix

\textbf{\large{Contents in Appendices:}}
\begin{itemize}
    \item In Appendix~\ref{appendix:agent_structure}, we provide detailed information of our agent architecture.
    \item In Appendix~\ref{appendix:implementation_details}, we provide implementation details.
    \item In Appendix~\ref{appendix:full_results}, we provide the full set of our experiment results.
\end{itemize}

\section{Details on Agent Structure}
\label{appendix:agent_structure}
In this section, we provide detailed information of our agent.
We will describe each of the modules shown in Figure~\ref{fig:pipeline}.
Some information here may be redundant with what we describe in Section~\ref{section:method}, we repeat them here for reader's convenience.

\subsection*{Notations}

We use \textit{game step} $t$ to denote one round of interaction between an agent with the \isquad environment.
We use $o_t$ to denote text observation at game step $t$, and $q$ to denote question text.
We use $L$ to refer to a linear transformation, superscript of $L$ denotes the activation function applied to the linear layer.
Brackets $[\cdot;\cdot]$ denote vector concatenation.

\subsection{Encoder}
\label{appendix:encoder}
At a game step $t$, the encoder takes text observation (a sentence) $o_t$, the question $q$, and the graph $G_t$ generated by the graph updater (if applicable) as input.
It first converts each input into vector representations, then aggregates them using attention mechanism.

\subsubsection{Text Encoder}
We use a transformer-based encoder, which consists of an embedding layer and a transformer block \citep{vaswani17transformer}.
Specifically, embeddings are initialized by vectors extracted from a BERT model \citep{devlin2018bert} that is pre-trained on large corpus and fine-tuned on SQuAD\footnote{We obtain the embeddings from HuggingFace (\url{https://huggingface.co/}), the BERT large model (uncased), whole word masking, fine-tuned on SQuAD.}.
The embedding size is 1024, they are fixed during training in all settings.

The transformer block consists of a stack of 4 convolutional layers, a self-attention layer, and a 2-layer MLP with a ReLU non-linear activation function in between. 
Within the block, each convolutional layer has 96 filters, with the kernel size of 7.
In the self-attention layer, we use a block hidden size $H$ of 96, as well as a single head attention mechanism.
Layer normalization \citep{ba16layernorm} is applied after each component inside the block. 
Following standard transformer training, we add positional embeddings into each block's input.

At every game step $t$, we use the same text encoder to process $o_t$ and $q$. 
The resulting representations are $h_{o_t} \in \mathbb{R}^{L_{o_t} \times H}$ and $h_{q} \in \mathbb{R}^{L_q \times H}$, where $L_{o_t}$ is the number of tokens in $o_t$, $L_q$ denotes the number of tokens in $q$, $H = 96$ is the hidden size.

\subsubsection{Graph Encoder}

We adopt the graph encoder from \citep{adhikari2020gata}, which is a model based on R-GCN \citep{schlichtkrull2018rgcn}.
Specifically, at the $l$-th layer of the R-GCN, for each node $i \in \mathcal{V}$, given the set of its neighbor nodes $\mathcal{V}^e_i \in \mathcal{V}$ under relation $e \in \mathcal{E}$, the R-GCN computes:
\begin{equation}
    \tilde{h}_i = \sigma\left(\sum_{e\in \mathcal{E}}\sum_{j\in \mathcal{V}^e_i} W^{l}_e [h^{l}_j; \text{Emb}_e] + W^{l}_0 [h^{l}_i; \text{Emb}_e]\right),
\end{equation}
where $W^{l}_e$ and $W^{l}_0$ are trainable parameters.
When the graph is discrete (i.e., word co-occurrence, relative position, SRL), we use ReLU as the activation function $\sigma$; when the graph is continuous (i.e., continuous belief), we use Tanh function as $\sigma$ to stabilize the model.

As the initial input $h^{0}$ to the graph encoder, we concatenate a node embedding vector and the averaged word embeddings of node text (e.g., a word in word co-occurrence graph, a chunk of a sentence in SRL graph).
Similarly, for each relation $e$, $\text{Emb}_e$ is the concatenation of a relation embedding vector and the averaged word embeddings of $e$'s label.
Both node embedding and relation embedding vectors are randomly initialized and trainable.

We utilize highway connections \citep{srivastava15highway} between layers:
\begin{equation}
\begin{aligned}
    g &= L^{\mathrm{Sigmoid}}(\tilde{h}_i), \\
    h^{l+1}_i &= g \odot \tilde{h}_i + (1 - g) \odot h^{l}_i, \\
\end{aligned}
\end{equation}
where $\odot$ indicates element-wise multiplication.

We use a 3-layer graph encoder, with a hidden size $H=96$ in each layer.
The node embedding size and relation embedding size are 100 and 32, respectively.
The number of bases we use is 3.
The final output of graph encoder is $h_{G_t} \in \mathbb{R}^{\mathcal{N} \times H}$, where $\mathcal{N}$ is the number of nodes in the graph.

\subsubsection{Attention}

\paragraph{Context-Query Attention}
To aggregate the question $q$ with context comes from various modality (i.e., text and graph), we adopt the context-query attention layer from \citep{yu18qanet}.
We use a unified notion $c$ to represent the context in the description of the context-query attention, the encoding of $c$ is denoted as $h_c \in \mathbb{R}^{L_c \times H}$.

The attention layer first uses two MLPs to convert both $h_c$ and $h_q$ into the same space, the resulting tensors are denoted as $h_c' \in \mathbb{R}^{L_c \times H}$ and $h_q' \in \mathbb{R}^{L_q \times H}$, in which $H = 96$.
Then, a tri-linear similarity function is used to compute the similarities between each pair of $h_c'$ and $h_q'$ items: 
\begin{equation}
    S = W[h_c'; h_q'; h_c' \odot h_q'],
\end{equation}
where $W$ is trainable parameters with hidden size 96.

Softmax of the resulting similarity matrix $S$ along both dimensions are computed, this produces $S^A$ and $S^B$. 
Information in the two representations are then aggregated by:
\begin{equation}
\begin{aligned}
    h_{cq} &= [h_c'; P; h_c'\odot P; h_c' \odot Q], \\
    P &= S_q h_q'^{\top}, \\
    Q &= S_q S_c^{\top} h_c'^{\top}. \\
\end{aligned}
\end{equation}
Next, a linear transformation projects the aggregated representations to a space with size $H=96$:
\begin{equation}
    h_{cq} = L(h_{cq}).
\end{equation}
Now, $h_{cq} \in \mathbb{R}^{L_c \times H}$ is aggregated context-query representation.

\paragraph{Context-Context Attention}
Given the aggregated text-query representations $h_{oq}$ and the aggregated graph-query representations $h_{gq}$, the context-context attention aims to merge them together and generate an overall representation that encodes all available information the agent has seen so far.

Specifically, the context-context attention is implemented as a stacked layers of transformer blocks.
The structure of these transformer blocks is similar with the text encoder blocks, except we append an extra attention layer after its self-attention mechanism.
This extra attention layer computes the attention between text-query representations with graph-query representations, followed with an extra layer normalization.
In each block, we use a stack of 2 convolutional layers, each convolutional layer has 94 filters with kernel size of 5.
It is worth noting that this additional attention layer is performed only when graph representations are enabled, and is skipped otherwise.
We stack 7 such transformer layers, they output $h_{og} \in \mathbb{R}^{L_{o_t} \times H}$, where $L_{o_t}$ is the length of $o_t$, $H=96$ is hidden size.

\subsubsection{Recurrent Component}
As mentioned in Section~\ref{section:exp_and_result}, we have a setting where the encoder is recurrent, so that the agent can incorporate history information into its representations.
In that specific setting, we use a GRU \citep{cho2014gru} as the recurrent component:
\begin{equation}
\begin{aligned}
    h_{\text{inp}} &= \textrm{MaskedMean}(h_{og}),\\
    h_t &= \mathrm{GRU}(h_{\text{inp}}, h_{t-1}),\\
\end{aligned}
\end{equation}
in which, the mean pooling is performed along the dimension of number of tokens, i.e., $h_{\text{inp}} \in \mathbb{R}^{H}$. $h_{t-1}$ is the output of the GRU cell at game step $t-1$.

\subsection{Action Generator}
Let $M \in \mathbb{R}^{H}$ denote the output of the attention layers described above:
\begin{equation}
    M = 
    \begin{cases}
        h_t  & \text{if recurrent,}\\
        \textrm{MaskedMean}(h_{og})  & \text{otherwise.}
    \end{cases}
\end{equation}

The action generator takes $M$ as input and generates rankings for all possible actions. 
As defined in \citep{yuan2020imrc}, a \cmd{\ctrlf} command is composed of two tokens (the token ``\cmd{\ctrlf}'' and the \query token).
Therefore, the action generator consists of three multi-layer perceptrons (MLPs): 
\begin{equation}
\begin{aligned}
    h_{\text{shared}} &= L_{\text{shared}}^{\text{ReLU}}(M), \\
    Q_{\text{action}} &= L_{\text{action}}(h_{\text{shared}}), \\
    h_{\text{query}} &= L^{\text{Tanh}}_{\text{query}}(h_{\text{shared}}),\\
    Q_{\text{query}} &= \textrm{Emb}(h_{\text{query}}). \\
\end{aligned}
\end{equation}
In which, $Q_{\text{action}}$ and $Q_{\text{query}}$ are Q-values of action token and \query token (when action token is ``\cmd{\ctrlf}''), respectively. 
The hidden size of $L_{\text{shared}}$ is 150.
The hidden size of $L_{\text{action}}$ is either 2 (hard, only \cmd{\ctrlf } and \cmd{stop} commands are allowed) or 4 (easy, \cmd{previous} and \cmd{next} are also allowed) depending on the game mode.
We follow \citep{press2016using}, tying the input embeddings and output embeddings.
Specifically, a linear layer $L_{\text{query}}$ followed by a Tanh activation projects $h_{\text{shared}}$ into the same space as the embeddings (with dimensionality of 1024), then the pre-trained BERT embedding matrix generates output logits $Q_{\text{query}}$ (Q-values) where the output size is same as the vocabulary size.

Under different settings where the selection spaces of \query are specified, we apply different masks to $Q_{\text{query}}$. 
For instance, in the setting where the \query is a word selected from $q + o_t$, we use a mask which has same size as the vocabulary, where only tokens appear in either $q$ and $o_t$ are set to 1.

\begin{figure*}[t!]
    \centering
    \includegraphics[width=0.9\textwidth]{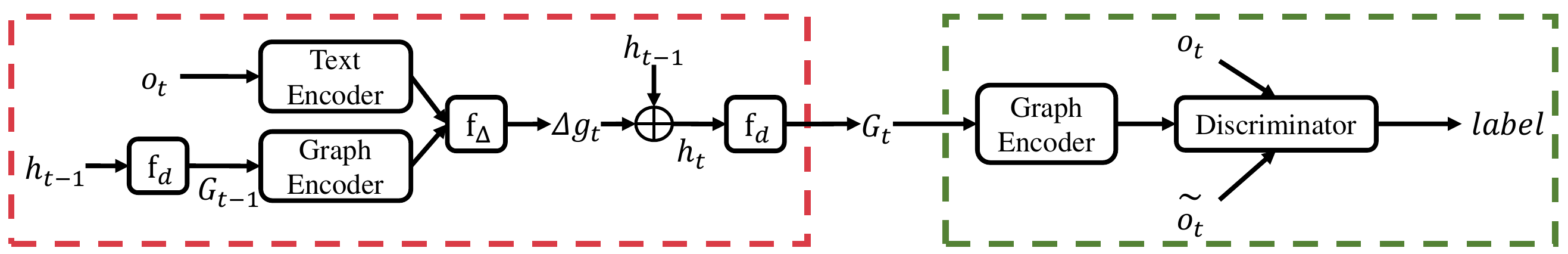}
    \caption{Graph updater for continuous belief graphs. }
    \label{fig:continuous_graph_updater}
\end{figure*}

\subsection{Question Answerer}

Whenever the action generator generates the command \cmd{stop}, or the agent has used up all its limit of moves, the information gathering phase terminates.
At this point, the agent has to use its current internal representations to answer the question.

The question answer is a simple MLP-based layer.
It takes $h_{og} \in \mathbb{R}^{L_{o_t} \times H}$ as input, and generates a head distribution and a tail distribution over tokens in $o_t$:
\begin{equation}
\begin{aligned}
    h_{\text{head}} &= L_0^\textrm{ReLU}(h_{og}), \\
    h_{\text{tail}} &= L_1^\textrm{ReLU}(h_{og}), \\
    p_{\text{head}} &= L_2^\textrm{Softmax}(h_{\text{head}}), \\
    p_{\text{tail}} &= L_3^\textrm{Softmax}(h_{\text{tail}}). \\
\end{aligned}
\end{equation}

\subsection{Graph Updater: Continuous Belief}
\label{appendix:gata_updater}

Among all the four proposed graph categories, only the continuous belief graph is generated by the agent, and the graph updater is trained with a data-driven approach.
Therefore, we describe the structure of this graph updater and the way we train it.
Because a large portion of this module is adopted from \citep{adhikari2020gata}, we provide a high level of the method and refer readers to \citep{adhikari2020gata} for detailed information.

We show the continuous belief graph updater training pipeline in Figure~\ref{fig:continuous_graph_updater}, it consists of two parts: the graph updater itself (red block on the left) and a decoder that helps to train the graph updater (green block on the right).
As mentioned in Section~\ref{subsection:gata}, the idea is to train a graph updater that can modify and maintain a graph $G_t$ using the text observation $o_t$ and the graph from previous game step $G_{t-1}$. 
The graph $G_t$ should contain sufficient information so that conditioned solely on $G_t$, a discriminator can differentiate true observation $o_t$ from negative sample $\tilde{o_t}$.

In Figure~\ref{fig:continuous_graph_updater}, text encoder and graph encoder are similar modules as described in Appendix~\ref{appendix:encoder}. 
The $\mathrm{f_\Delta}$ function is a layer with attention mechanism inside, it aggregates the text representations and graph representations.
The $\mathrm{f_\Delta}$ function outputs a vector $\Delta g_t$, which represents the new information seen in the new observation $o_t$, compared to the graph at previous game step $G_{t-1}$.

The $\oplus$ function is a graph operation function that produces the new belief representation $h_t$ given $h_{t-1}$ and $\Delta g_t$:
\begin{equation}
    h_t = h_{t-1} \oplus \Delta g_t.
\end{equation}
The graph operation function is implemented with a GRU \citep{cho2014gru}.
The function $\mathrm{f_d}$ is an MLP that decodes the recurrent state $h_t$ into a real-valued adjacency tensor (i.e., the continuous belief graph $\mathcal{G}_{t}$).

At the decoder side (green block on the right), the graph representations of $G_t$ are concatenated with both the text representations of $o_t$ and $\tilde{o_t}$, the resulting vectors are fed into an MLP-based discriminator.
The discriminator is trained with the standard binary cross-entropy (BCE) loss.

After pre-training, the graph updater (red block on the left) is fixed and plugged into our RL agent to produce continuous belief graphs.

\section{Implementation Details}
\label{appendix:implementation_details}

In this section, we provide hyperparameters and other implementation details.

For our full agent, we adopt the training procedure of DRQN \citep{hausknecht2015deep,yuan2018counting} to train the agent.
For the agent variants where recurrent component is absent, we use the training procedure of DQN \citep{mnih2013playing}.

For all experiments, we use \emph{Adam} \citep{kingma14adam} as the optimizer.
The learning rate is set to 0.00025 with a clip gradient norm of 5.

We use a prioritized replay buffer with memory size of 500,000, and a priority fraction of 0.5.
During model update, we use a replay batch size of 64.
We use a discount $\gamma = 0.9$.
We use noisy nets, with a $\sigma_0$ of 0.5.
We update target network after every 1000 episodes.
In DQN training, we sample the multi-step return $n \sim \text{Uniform}[1, 3]$.
In DRQN training, every sampled data point in the replay batch is a sequence of 2 consecutive transitions. 
We use the first transition to estimate the recurrent states, and the second for updating the model parameters.
We refer readers to \citet{hessel18rainbow} for more information about different components of DQN training.

We train all agents with 3 different random seeds. 
We choose the random seed which produces the best validation performance, and report its scores on the test set.
To be comparable with \imrc, we also train our agents with 1 million episodes, each episode has a maximum number of steps 20.

We train all agents for 1 million episodes, this is the same number of episodes reported in \citep{yuan2020imrc}.
Running speed of agents depend on the specific configuration, e.g., the type of graph equipped by an agent.
On average, achieving best validation score takes an agent about 3 days on a single Nvidia P100 GPU.

\section{Full Results}
\label{appendix:full_results}

In Table~\ref{tab:result_table_full_slot_1},\ref{tab:result_table_full_slot_3},\ref{tab:result_table_full_slot_5}, we provide full results on our experiments.
Although the only metric to evaluate an agent's performance on the \imrc task is the \fone score between the prediction with the ground-truth answers, in \citep{yuan2020imrc}, the authors also monitor agents' sufficient information reward.
Specifically, sufficient information rewards are binary rewards representing if the final observation (either the agent generates the \cmd{stop} action, or it has used up all its moves) contains the ground-truth answer as a sub-string.
Intuitively, because of the extractive nature of the question answerer module, the agent can answer the question correctly if and only if it achieves a 1.0 sufficient information reward on a specific data point. 
We provide the sufficient information rewards of our agents in the full result tables, colored in blue.

\begin{table*}
    \centering
    \scriptsize
    \begin{tabular}{c|ccc|ccc|c}
        \toprule
        & \multicolumn{3}{c|}{Easy Mode} & \multicolumn{3}{c|}{Hard Mode} &  \\
        Agent & $q$ & $q + o_t$ & vocab & $q$ & $q + o_t$ & vocab & \%RI\\
        \midrule %
        \imrc \citep{yuan2020imrc}      &  0.575 \color{blue}(0.747) & 0.579 \color{blue}(0.739) & 0.583 \color{blue}(0.753) & 0.524 \color{blue}(0.684) & 0.357 \color{blue}(0.477) & 0.264 \color{blue}(0.363) & -- \\
        \midrule
        Ours (co-occur)                         & 0.632 \color{blue}(0.779) & 0.624 \color{blue}(0.763) & 0.635 \color{blue}(0.770) & 0.582 \color{blue}(0.724) & 0.426 \color{blue}(0.533) & 0.258 \color{blue}(0.338) & 9.16 \color{blue}(3.40) \\
        Ours (rel. pos.)                        & 0.634 \color{blue}(0.776) & 0.634 \color{blue}(0.774) & 0.642 \color{blue}(0.779) & 0.562 \color{blue}(0.696) & 0.440 \color{blue}(0.553) & 0.250 \color{blue}(0.338) & 9.18 \color{blue}(3.81) \\
        Ours (SRL)                              & 0.616 \color{blue}(0.751) & 0.641 \color{blue}(0.779) & 0.638 \color{blue}(0.782) & 0.603 \color{blue}(0.740) & 0.434 \color{blue}(0.538) & 0.253 \color{blue}(0.338) & 9.98 \color{blue}(3.99) \\
        Ours (cont.)                            & 0.617 \color{blue}(0.757) & 0.628 \color{blue}(0.763) & 0.616 \color{blue}(0.758) & 0.597 \color{blue}(0.744) & 0.436 \color{blue}(0.542) & 0.257 \color{blue}(0.338) & 9.14 \color{blue}(3.46) \\
        Ours (\textcolor{darkgreen}{ensemble})  & 0.677 \color{blue}(0.789) & 0.691 \color{blue}(0.799) & 0.686 \color{blue}(0.795) & 0.627 \color{blue}(0.735) & 0.472 \color{blue}(0.555) & 0.276 \color{blue}(0.338) & 18.53 \color{blue}(6.04) \\
        \midrule
        Ours (co-occur) w/o RNN                         & 0.572 \color{blue}(0.708) & 0.613 \color{blue}(0.753) & 0.607 \color{blue}(0.744) & 0.556 \color{blue}(0.689) & 0.397 \color{blue}(0.495) & 0.269 \color{blue}(0.359) & 4.73 \color{blue}(-0.20) \\
        Ours (rel. pos.) w/o RNN                        & 0.574 \color{blue}(0.702) & 0.621 \color{blue}(0.761) & 0.608 \color{blue}(0.746) & 0.540 \color{blue}(0.676) & 0.409 \color{blue}(0.504) & 0.255 \color{blue}(0.338) & 4.27 \color{blue}(-1.05) \\
        Ours (SRL) w/o RNN                              & 0.559 \color{blue}(0.703) & 0.628 \color{blue}(0.772) & 0.631 \color{blue}(0.773) & 0.540 \color{blue}(0.672) & 0.411 \color{blue}(0.511) & 0.266 \color{blue}(0.347) & 5.50 \color{blue}(0.38) \\
        Ours (cont.) w/o RNN                            & 0.588 \color{blue}(0.724) & 0.604 \color{blue}(0.739) & 0.605 \color{blue}(0.747) & 0.561 \color{blue}(0.693) & 0.414 \color{blue}(0.513) & 0.258 \color{blue}(0.338) & 5.23 \color{blue}(-0.30) \\
        Ours (\textcolor{darkgreen}{ensemble}) w/o RNN  & 0.598 \color{blue}(0.701) & 0.659 \color{blue}(0.769) & 0.664 \color{blue}(0.773) & 0.593 \color{blue}(0.695) & 0.441 \color{blue}(0.526) & 0.278 \color{blue}(0.338) & 12.29 \color{blue}(0.93) \\
        \bottomrule
    \end{tabular}
    \caption{\#Memory slot = 1. Testing \fone in \textbf{black} and sufficient information rewards in \textcolor{blue}{\textbf{blue}}. \%RI represents relative improvement over \imrc on corresponding metric, across settings.}
    \label{tab:result_table_full_slot_1}
\end{table*}

\begin{table*}
    \centering
    \scriptsize
    \begin{tabular}{c|ccc|ccc|c}
        \toprule
        & \multicolumn{3}{c|}{Easy Mode} & \multicolumn{3}{c|}{Hard Mode} &  \\
        Agent & $q$ & $q + o_t$ & vocab & $q$ & $q + o_t$ & vocab & \%RI\\
        \midrule %
        \imrc \citep{yuan2020imrc}      & 0.637 \color{blue}(0.738) & 0.651 \color{blue}(0.734) & 0.624 \color{blue}(0.738) & 0.524 \color{blue}(0.740) & 0.362 \color{blue}(0.729) & 0.261 \color{blue}(0.719) & -- \\
        \midrule
        Ours (co-occur)                         & 0.674 \color{blue}(0.863) & 0.665 \color{blue}(0.859) & 0.675 \color{blue}(0.872) & 0.605 \color{blue}(0.780) & 0.446 \color{blue}(0.585) & 0.260 \color{blue}(0.338) & 9.06 \color{blue}(3.52) \\
        Ours (rel. pos.)                        & 0.677 \color{blue}(0.874) & 0.665 \color{blue}(0.850) & 0.664 \color{blue}(0.877) & 0.615 \color{blue}(0.789) & 0.438 \color{blue}(0.575) & 0.257 \color{blue}(0.338) & 8.66 \color{blue}(3.54) \\
        Ours (SRL)                              & 0.681 \color{blue}(0.896) & 0.678 \color{blue}(0.887) & 0.654 \color{blue}(0.841) & 0.600 \color{blue}(0.780) & 0.440 \color{blue}(0.564) & 0.258 \color{blue}(0.338) & 8.45 \color{blue}(3.33) \\
        Ours (cont.)                            & 0.676 \color{blue}(0.870) & 0.642 \color{blue}(0.850) & 0.662 \color{blue}(0.864) & 0.592 \color{blue}(0.746) & 0.426 \color{blue}(0.544) & 0.282 \color{blue}(0.389) & 8.26 \color{blue}(3.49) \\
        Ours (\textcolor{darkgreen}{ensemble})  & 0.714 \color{blue}(0.864) & 0.713 \color{blue}(0.867) & 0.701 \color{blue}(0.851) & 0.650 \color{blue}(0.783) & 0.471 \color{blue}(0.567) & 0.278 \color{blue}(0.338) & 15.80 \color{blue}(2.75) \\
        \midrule
        Ours (co-occur) w/o RNN                         & 0.624 \color{blue}(0.810) & 0.666 \color{blue}(0.870) & 0.658 \color{blue}(0.860) & 0.577 \color{blue}(0.732) & 0.417 \color{blue}(0.523) & 0.272 \color{blue}(0.362) & 5.89 \color{blue}(0.34) \\
        Ours (rel. pos.) w/o RNN                        & 0.625 \color{blue}(0.840) & 0.650 \color{blue}(0.838) & 0.649 \color{blue}(0.835) & 0.577 \color{blue}(0.741) & 0.420 \color{blue}(0.524) & 0.305 \color{blue}(0.386) & 7.48 \color{blue}(1.19) \\
        Ours (SRL) w/o RNN                              & 0.593 \color{blue}(0.772) & 0.639 \color{blue}(0.865) & 0.636 \color{blue}(0.828) & 0.568 \color{blue}(0.732) & 0.400 \color{blue}(0.510) & 0.275 \color{blue}(0.371) & 2.91 \color{blue}(-1.18) \\
        Ours (cont.) w/o RNN                            & 0.641 \color{blue}(0.820) & 0.645 \color{blue}(0.840) & 0.667 \color{blue}(0.854) & 0.576 \color{blue}(0.742) & 0.420 \color{blue}(0.537) & 0.261 \color{blue}(0.338) & 5.43 \color{blue}(-0.57) \\
        Ours (\textcolor{darkgreen}{ensemble}) w/o RNN  & 0.663 \color{blue}(0.803) & 0.704 \color{blue}(0.859) & 0.709 \color{blue}(0.858) & 0.627 \color{blue}(0.748) & 0.460 \color{blue}(0.538) & 0.293 \color{blue}(0.350) & 14.15 \color{blue}(0.31) \\
        \bottomrule
    \end{tabular}
    \caption{\#Memory slot = 3. Testing \fone in \textbf{black} and sufficient information rewards in \textcolor{blue}{\textbf{blue}}. \%RI represents relative improvement over \imrc on corresponding metric, across settings.}
    \label{tab:result_table_full_slot_3}
\end{table*}

\begin{table*}
    \centering
    \scriptsize
    \begin{tabular}{c|ccc|ccc|c}
        \toprule
        & \multicolumn{3}{c|}{Easy Mode} & \multicolumn{3}{c|}{Hard Mode} &  \\
        Agent & $q$ & $q + o_t$ & vocab & $q$ & $q + o_t$ & vocab & \%RI\\
        \midrule %
        \imrc \citep{yuan2020imrc}      & 0.666 \color{blue}(0.716) & 0.656 \color{blue}(0.706) & 0.661 \color{blue}(0.731) & 0.551 \color{blue}(0.739) & 0.364 \color{blue}(0.733) & 0.218 \color{blue}(0.713) & -- \\
        \midrule
        Ours (co-occur)                         & 0.680 \color{blue}(0.883) & 0.670 \color{blue}(0.905) & 0.665 \color{blue}(0.879) & 0.628 \color{blue}(0.804) & 0.444 \color{blue}(0.572) & 0.258 \color{blue}(0.338) & 9.84 \color{blue}(3.84) \\
        Ours (rel. pos.)                        & 0.686 \color{blue}(0.897) & 0.677 \color{blue}(0.900) & 0.665 \color{blue}(0.898) & 0.622 \color{blue}(0.802) & 0.446 \color{blue}(0.583) & 0.253 \color{blue}(0.338) & 9.76 \color{blue}(4.66) \\
        Ours (SRL)                              & 0.675 \color{blue}(0.884) & 0.680 \color{blue}(0.902) & 0.680 \color{blue}(0.920) & 0.609 \color{blue}(0.767) & 0.441 \color{blue}(0.563) & 0.257 \color{blue}(0.338) & 9.56 \color{blue}(3.43) \\
        Ours (cont.)                            & 0.699 \color{blue}(0.916) & 0.693 \color{blue}(0.914) & 0.696 \color{blue}(0.925) & 0.629 \color{blue}(0.805) & 0.455 \color{blue}(0.595) & 0.257 \color{blue}(0.338) & 12.19 \color{blue}(6.25) \\
        Ours (\textcolor{darkgreen}{ensemble})  & 0.736 \color{blue}(0.895) & 0.733 \color{blue}(0.908) & 0.725 \color{blue}(0.898) & 0.665 \color{blue}(0.798) & 0.484 \color{blue}(0.582) & 0.277 \color{blue}(0.338) & 18.79 \color{blue}(4.64) \\
        \midrule
        Ours (co-occur) w/o RNN                         & 0.631 \color{blue}(0.814) & 0.674 \color{blue}(0.896) & 0.682 \color{blue}(0.918) & 0.586 \color{blue}(0.753) & 0.412 \color{blue}(0.513) & 0.259 \color{blue}(0.338) & 6.51 \color{blue}(0.03) \\
        Ours (rel. pos.) w/o RNN                        & 0.636 \color{blue}(0.833) & 0.682 \color{blue}(0.908) & 0.669 \color{blue}(0.893) & 0.572 \color{blue}(0.734) & 0.412 \color{blue}(0.512) & 0.258 \color{blue}(0.338) & 5.97 \color{blue}(-0.29) \\
        Ours (SRL) w/o RNN                              & 0.619 \color{blue}(0.793) & 0.683 \color{blue}(0.907) & 0.635 \color{blue}(0.822) & 0.571 \color{blue}(0.727) & 0.434 \color{blue}(0.554) & 0.265 \color{blue}(0.345) & 6.29 \color{blue}(-0.72) \\
        Ours (cont.) w/o RNN                            & 0.638 \color{blue}(0.836) & 0.664 \color{blue}(0.884) & 0.653 \color{blue}(0.858) & 0.575 \color{blue}(0.731) & 0.427 \color{blue}(0.533) & 0.259 \color{blue}(0.338) & 6.02 \color{blue}(-0.72) \\
        Ours (\textcolor{darkgreen}{ensemble}) w/o RNN  & 0.668 \color{blue}(0.803) & 0.733 \color{blue}(0.905) & 0.715 \color{blue}(0.879) & 0.624 \color{blue}(0.742) & 0.466 \color{blue}(0.546) & 0.280 \color{blue}(0.338) & 14.99 \color{blue}(0.16) \\
        \bottomrule
    \end{tabular}
    \caption{\#Memory slot = 5. Testing \fone in \textbf{black} and sufficient information rewards in \textcolor{blue}{\textbf{blue}}. \%RI represents relative improvement over \imrc on corresponding metric, across settings.}
    \label{tab:result_table_full_slot_5}
\end{table*}

\end{document}